\pdfoutput=1
\documentclass[11pt]{article}

\usepackage{acl}

\usepackage{times}
\usepackage{latexsym}
\usepackage{smile}
\usepackage{amsfonts}
\usepackage{amssymb}
\usepackage[T1]{fontenc}
\usepackage[utf8]{inputenc}

\usepackage{microtype}

\usepackage{xcolor, colortbl}  
\usepackage{inconsolata}
\usepackage{multirow}
\usepackage{subfigure}
\usepackage{algorithm}
\usepackage{algorithmic}
\usepackage{listings}
\usepackage{pythonhighlight}
\usepackage{fancyvrb}
\usepackage{xspace}

\RecustomVerbatimCommand{\VerbatimInput}{VerbatimInput}{fontsize=\footnotesize,
 frame=single,  
 framesep=0.5em, % separation between frame and text
 labelposition=topline,
}
\usepackage{tcolorbox}
\tcbuselibrary{listings}
\newcommand{\ours}{\textsc{Ram-EHR}}
\newcommand{\dataset}{\textsc{Cradle}}

\definecolor{nblue}{cmyk}{0.95,0.0,0.2,0.2}
\newcommand{\blue}[1]{\textcolor{nblue}{\small #1}}

\title{\ours{}: Retrieval Augmentation Meets Clinical Predictions \\ on Electronic Health Records}

\author{Ran Xu$^1$\thanks{~Equal contribution.},  Wenqi Shi$^2$\footnotemark[1], Yue Yu$^2$, Yuchen Zhuang$^2$, Bowen Jin$^3$ \\ \bf May D. Wang$^2$, Joyce C. Ho$^1$, Carl Yang$^1$ \\  
$^1$ Emory University  \quad  $^2$ Georgia Institute of Technology  \\  $^3$ University of  Illinois at Urbana Champaign \\
\texttt{\{ran.xu,joyce.c.ho,j.carlyang\}@emory.edu}, \\
\texttt{\{wshi83,yueyu,yczhuang,maywang\}@gatech.edu}, \texttt{ bowenj4@illinois.edu}
}

\begin{document}
\maketitle
\begin{abstract}
We present \ours{}, a {R}etrieval {A}ug{M}entation pipeline to improve clinical predictions on {E}lectronic {H}ealth {R}ecords (EHRs).
\ours{} first collects multiple knowledge sources, converts them into text format, and uses dense retrieval to obtain information related to medical concepts. 
This strategy addresses the difficulties associated with complex names for the concepts.
\ours{} then augments the local EHR predictive model co-trained with consistency regularization to capture complementary information from patient visits and summarized knowledge.
% We then design an additional module to augment the local EHR predictive model, which is co-trained with consistency regularization to capture complementary information from patient visits and summarized knowledge. 
Experiments on two EHR datasets show the efficacy of \ours{} over previous knowledge-enhanced baselines (3.4\% gain in AUROC and 7.2\% gain in AUPR), emphasizing the effectiveness of the summarized knowledge from \ours{} for clinical prediction tasks. The code will be published at \url{https://github.com/ritaranx/RAM-EHR}.

\end{abstract}
% \vspace{-1ex}
% \vspace{-1ex}
\section{Introduction}
% \vspace{-0.5ex}
\label{sec:intro}
Electronic Health Records (EHRs), encompassing detailed information about patients such as symptoms, diagnosis, and medication, are widely used by physicians to 
% encode patient information. 
deliver patient care~\citep{jensen2012mining,cowie2017electronic,shi2024ehragent}. 
% An important area of application for EHR is \emph{predictive healthcare}, where
Recently, a vast amount of deep learning models have been developed on EHR data~\citep{choi2016retain,choi2020learning,gao2020stagenet,pmlr-v193-xu22a,cai2022hypergraph,wang-etal-2023-hierarchical} for various downstream prediction tasks (e.g., disease diagnosis, risk prediction) to facilitate precision healthcare. 

To further improve the downstream predictive performance, several works attempt to augment the EHR visits with external knowledge. 
% To enhance predictive accuracy, several studies have integrated external knowledge with EHR visits for downstream predictive tasks.
For example, \citet{van-aken-etal-2021-clinical} and \citet{naik-etal-2022-literature} incorporate additional clinical notes, although these clinical notes can be noisy and contain {irrelevant contents} for clinical predictions; another solution is to leverage external clinical knowledge graphs (KGs), such as UMLS~\citep{chandak2023building}, 
which contains rich medical concepts (e.g., disease, medications) and their corresponding relationships~\citep{jin2023large}.
% which represent medical concepts (e.g., disease, medications, procedures) and their relationships in a structured way. 
% When combining KG with EHR visits, the semantic knowledge in KG can often complement EHR modeling with improved performance
Integrating KGs with EHRs has been shown to boost model performance~\citep{choi2017gram,xu2023seqcare,gao2023leveraging}.
However, these works mostly rely on knowledge from a single source and medical KGs mainly focus on specific types of relations (e.g., hierarchical relations), which do not comprehensively capture the semantic information for medical codes (e.g., phenotype). 
% Besides, due to the 
Besides, it is non-trivial to align medical codes in EHRs with KGs due to the non-uniformity of surface names (e.g., abbreviations or colloquial terms)~\citep{hao2021medto,zhang-etal-2022-knowledge}.
% \citet{ye2021medretriever} extracts medical texts using string similarity, which can still be imprecise due to the complex surface names of medical codes~\cite{zhu-etal-2023-controllable}.
% directly aligning medical codes in EHR with KGs is challenging due to their complex surface names, often causing entity linking issues~\cite{zhu-etal-2023-controllable}.
% Besides, matching medical codes from EHRs directly with KG is challenging due to the complex surface names for medical codes, which often leads to entity linking issues~\cite{zhu-etal-2023-controllable}.
\citet{jiang2023graphcare} use knowledge generated from large language models (LLMs) to assist EHR prediction, but LLMs 
% often suffer from the hallucination issue~\citep{manakul-etal-2023-selfcheckgpt} and 
may not always provide the most relevant knowledge for target tasks and face the risk of hallucination.
Effectively leveraging external knowledge to facilitate EHR predictive tasks remains a significant challenge.
% How to effectively leverage external knowledge to assist EHR predictive tasks is still challenging.
% enabling effective learning of 

In this work, we propose {\ours}, a retrieval-augmented framework tailored for clinical predictive tasks on EHRs. 
Instead of leveraging a single knowledge source, {\ours} collects
multiple knowledge sources (e.g., KGs, scientific literature) and converts them to text corpus, which enjoys the merits of a more comprehensive coverage of knowledge in a unified format. 
Then, to obtain unified representations for different knowledge sources, we leverage dense retrieval (DR) \citep{karpukhin-etal-2020-dense,izacard2022unsupervised,yu-etal-2022-coco,lin-etal-2023-train} to encode corpus and medical codes as dense vectors, intuitively capturing the semantics of medical codes and addressing the alignment issue between EHR and external knowledge.
% complexity associated with names of medical codes. 
Finally, to reduce irrelevant information and reduce the length of the input text, we utilize an LLM to summarize the top-retrieved passages into concise and informative knowledge summaries relevant to downstream tasks for each medical code. 
This process enhances the relevance and utility of the retrieved knowledge for clinical tasks.
% and customizes it for clinical tasks.

% To effectively leverage external knowledge to assist clinical prediction, we augment the local EHR predictive model which uses solely patient visit information, where summarized passages and medical codes are concatenated and fed into a moderate-size, pre-trained language model.
To leverage external knowledge to assist clinical prediction, we introduce a retrieval-augmented model alongside the local EHR predictive model, which relies solely on patient visit information. 
The augmented model concatenates summarized passages and medical codes, feeding them into a moderate-size, pre-trained language model.
We then co-train the local model and the augmented model with a consistency regularization, which captures the \emph{complementary information} from patient visits and summarized knowledge and helps the model with better generalization~\citep{wan2009co,pmlr-v162-lang22a}.

We verify the effectiveness of \ours{} by conducting experiments on two EHR datasets and show that \ours{} outperforms strong knowledge-enhanced predictive baselines 
by 3.4\% in AUROC and 7.2\% in AUPR on average. 
Our analysis further confirms the advantage of leveraging multi-source external knowledge as well as retrieval augmentation as plugins to assist vanilla EHR predictive models based on visits only.

Our contribution can be summarized as 
\begin{itemize}
    \item We introduce \ours{}, an innovative retrieval-augmented framework designed to harness external knowledge to enhance EHR-based clinical predictions. Notably, \ours{} offers flexibility and can seamlessly integrate diverse sources of knowledge. 
    % Additionally, we employ an LLM to generate summarized knowledge for medical codes to further enhance the framework's capabilities.
    \item We introduce a co-training approach aimed at effectively leveraging the complementary information derived from summarized knowledge and patient visit records.
    \item We conduct experiments on two EHR datasets and observe that \ours{} leads to consistent gains when compared with other knowledge-enhanced EHR predictive models. We also demonstrate that \ours{} can serve as a generic plugin for multiple types of base models. Additional human studies justify the usefulness of summarized knowledge for assisting clinical prediction tasks.
\end{itemize}
% for tackling the limited coverage issue of label names as well as the efficacy of multi-stage training for improving performance

% Finally, most relevant passages and questions are linearly concatenated and fed into pre-trained language model for question answering. 

% gao2023leveraging
% \citep{rosenthal-etal-2019-leveraging}
% \citep{

% \cite{lin-etal-2023-train}

% Introductions to the problem, existing works and challenges

% Transformer~\cite{} 

% TransEHR~\cite{transehr}

% \cite{choi2020learning}

% HyperGraph \cite{}
% EHR + KG

% \cite{gao2023leveraging}

% \cite{jiang2023graphcare}
% \cite{xu2023seqcare}
% Clinical Notes:

% \cite{van-aken-etal-2021-clinical}

% \cite{naik-etal-2022-literature}

\section{Related Works}
\label{sec:related}
\paragraph{Retrieval Augmented Learning and its Application in Clinical Domain.}
Retrieval augmented learning, which collects additional contextual information from external corpus, has shown effectiveness on diverse tasks including language modeling~\citep{pmlr-v162-borgeaud22a,wang-etal-2023-shall}, knowledge-intensive NLP~\citep{lewis2020retrieval,shi2023replug}, commonsense reasoning~\citep{yu-etal-2022-retrieval}, 
code genereation~\citep{parvez-etal-2021-retrieval-augmented}, and few/zero-shot learning~\citep{xu2023weakly,izacard2022few}. 
Compared to the general domain, the application of retrieval-augmented learning to clinical tasks is still under-explored. 
Some efforts have been paid to retrieval augmented clinical language models~\citep{zakka2024almanac}, clinical decision making~\citep{shi2023retrieval,thompson2023large} and  clinical question answering~\citep{wang2023augmenting,xiong2024benchmarking,jeong2024improving}.
The most relevant works are \citet{ye2021medretriever,naik-etal-2022-literature}, which leverage clinical literature to augment clinical predictive models. Compared to these works, our contribution lies in two folds: (1) we design a retrieval augmentation for \emph{structured} EHRs with a diverse collection of external knowledge, which provides more relevant information for target clinical prediction tasks;
(2) we incorporate a co-training scheme to leverage both the visit-level information and external knowledge for predictions.

\paragraph{Knowledge-enhanced EHR Predictive Models.}
% Most of the existing approaches focus on leveraging \emph{knowledge graphs} as external sources to clinical prediction tasks. 

Many studies attempt to harness external knowledge for clinical prediction tasks. 
The majority of them leverage structured knowledge, such as medical ontology \citep{choi2017gram,panigutti2020doctor,lu2021collaborative}, to capture hierarchical relationships among medical codes, or employ personalized knowledge graphs \citep{xu2023seqcare,jiang2023graphcare} to integrate patient-specific information. However, these methods often suffer from limited coverage of all medical codes due to the complexity of surface names. Alternatively, some approaches utilize unstructured medical text for health prediction tasks \citep{ye2021medretriever}. 
However, \citet{ye2021medretriever} rely on a restricted corpus of approximately 30,000 passages as their external corpus, resulting in limited coverage. 
Besides, it only considers ICD codes and uses the Levenshtein distance for retrieval, which cannot fully capture the semantics of medical codes.

% \citep{zhang-etal-2022-pm2f2n}

% \vspace{-0.8ex}
\section{Methodology}
\label{sec:method}
% \vspace{-0.6ex}
\begin{figure*}[t]
    \centering    
    \includegraphics[width=0.99\linewidth]{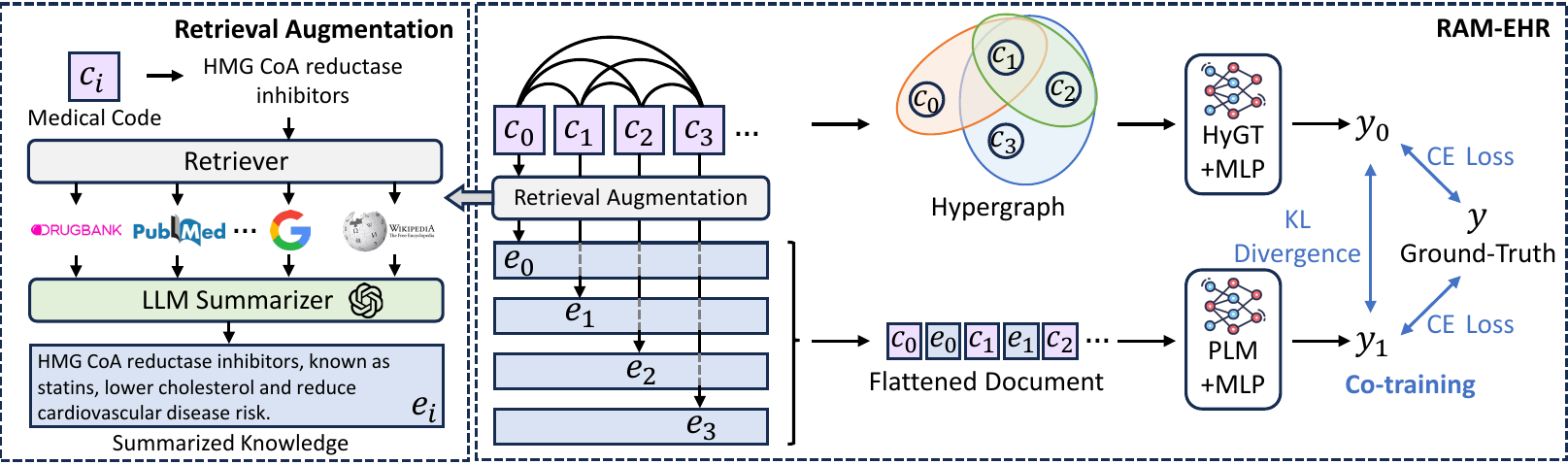}
    % \vspace{-1ex}
    \caption{An overview of retrieval augmentation framework (left) and a detailed workflow of {\ours{}} (right). \ours{} initially gathers multiple knowledge sources and converts them into textual format. We then use dense retrieval to obtain information related to medical concepts. Next, we design an additional module to augment the local EHR predictive model co-trained with consistency regularization, capturing complementary information from both patient visits and summarized knowledge.}
    % \vspace{-ex}
    \label{fig:overview}
\end{figure*}

Figure~\ref{fig:overview} presents a workflow of \ours{}, with a specific focus on dense retrieval from multiple knowledge sources and consistency regularization with co-training schema. The details of \ours{} will be introduced in the following sections.

\subsection{Problem Setup}
% \vspace{-0.3ex}
The EHR data consists of a group of patients $\cP$ with corresponding hospital visits $V=\{v_1, v_2, ..., v_{|V|}\}$. Each visit $v_i$ includes a set of medical codes $C_i \subset \mathcal{C}$, where $\cC$ is the total set of medical codes for $\cP$.
In this study, $\cC$ contains multiple types of medical codes including \emph{diseases}, \emph{medications}, and \emph{procedures}. Each medical code $c_i \in C_i$ is a \emph{clinical concept}, and it is associated with a name $s_i$ in the form of \emph{short text snippets}.
Given the clinical record $v_i$ with the involved medical codes $C_i$, we aim to predict the patient's clinical outcome $y_i$ (a binary label).

% For patient $p$ in patient group $\cP$, the health record $V$ for $p$ contains visits $V=\{v_1, v_2, ..., v_{t}\}$, where $t$ is the number of visits. Each visit $v_i$ includes a set of medical codes $C_i \subset \mathcal{C}$, where $\cC$ is the total set of medical codes for $\cP$. 
% In this study, $\cC$ contains multiple types of medical codes including \emph{diseases}, \emph{medications}, and \emph{procedures}. 
% Each medical code is a clinical concept associated with a name in short text snippets.
% Each medical code $c_i$ is a \emph{clinical concept}, and it is associated with a name $s_i$ in the form of \emph{short text snippets}.
% Given the clinical record $V$, we aim to predict the patient's clinical outcome $y$ (a binary label). \ran{avoid multiple visits}

\vspace{-1ex}
\subsection{Retrieval Augmentation w/ Medical Codes}
Existing approaches often treat each visit as context-free vectors, which fail to capture the concrete semantics of medical codes. 
Being aware of this, we aim to create the summarized knowledge for each medical code $c_i$ using its surface name $s_i$ via retrieval augmentation with additional contexts. 
% The details are described as follows.

\noindent \textbf{Multi-source Corpus Creation}.
Retrieval augmentation requires additional corpora as external knowledge. 
To ensure the coverage of clinical knowledge, we collect a diverse external resources $\cM=\{d_1, d_2, \ldots, d_{|\cM|}\}$. 
% For instance, $\cM$ can comprise a general encyclopedia, a knowledge graph, or domain-specific literature. 
We represent each knowledge unit as a raw text to facilitate retrieval, described as follows:
\begin{itemize}
    \item \textbf{PubMed}\footnote{\url{https://pubmed.ncbi.nlm.nih.gov/}}: PubMed is a free search engine accessing primarily the MEDLINE database of references and abstracts on life sciences and biomedical topics. It provides users with access to millions of scientific documents, including research papers, reviews, and other scholarly articles. We use the \texttt{Entrez} package to extract the PubMed articles\footnote{\url{https://biopython.org/docs/1.75/api/Bio.Entrez.html}}, resulting in 230k documents.
    \item \textbf{DrugBank}\footnote{\url{https://go.drugbank.com/releases/latest}}~\citep{wishart2008drugbank}: DrugBank is a comprehensive and freely accessible online database containing information on drugs and drug targets. It integrates detailed drug data (chemical, pharmacological, and pharmaceutical) with comprehensive information on drug targets (sequence, structure, and pathway). We use the data from the original database, which contains 355k documents.
    \item \textbf{Medical Subject Headings (MeSH)}\footnote{\url{https://www.ncbi.nlm.nih.gov/mesh/}}: Medical Subject Headings (MeSH) is a comprehensive controlled collection for indexing journal articles and books in the life sciences. It organizes information on biomedical and health-related topics into a hierarchical structure. The corpus contains 32.5k documents covering various medical concepts.
    \item \textbf{Wikipedia}\footnote{\url{https://www.wikipedia.org/}}~\citep{vrandevcic2014wikidata}: Wikipedia is a free, web-based, collaborative, multilingual encyclopedia project that is supported by the non-profit Wikimedia Foundation. We extract web pages that contain medical-related information by using the medical codes list (e.g., ICD10 and ATC), resulting in 150k documents.
    \item \textbf{KG}\footnote{\url{https://github.com/mims-harvard/PrimeKG}}~\citep{chandak2023building}: We use PrimeKG in our experiments. It offers a comprehensive overview of diseases, medications, side effects, and proteins by merging 20 biomedical sources to detail 17,080 diseases across ten biological levels. For this study, we select knowledge triplets that contain medical codes within three types (disease, medication, procedure) used in this work, resulting in 707k triplets. We use the template in Appendix~\ref{apd:format} to transform these triplets into sentences.
\end{itemize}
% Pubmed, KG, Drugbank, MeSH, Wikipedia

\noindent \textbf{Passage Retrieval}.
Given a collection of $|\cM|$ passages, the objective of the retriever is to transform passages in a \emph{dense} vector, so that it can efficiently retrieve the most relevant information to the input query. 
In our work, we adopt Dragon~\citep{lin-etal-2023-train}, a dual-encoder model with strong performance across domains as the retriever. 
% For each medical code $c_i$ with its name $s_i$, we estimate the relevance between the name and passages in $\cM$ by computing a single similarity score between their \texttt{[CLS]} token representations. 
Specifically, we first use the passage encoder $R_{D}(\cdot)$ to build an index for corpus $\cM$ to support retrieval.
Then, at runtime, we use the query encoder $R_Q(\cdot)$ to map the input to an embedding (same dimension as the passage embedding) and calculate the similarity as
% \begin{equation}
% \setlength{\abovedisplayskip}{4pt}
% \setlength{\belowdisplayskip}{4pt}
$f(q, d)={R}_{Q}(q)^{\top}{R}_{D}(d)$.
% \label{eq:sim}
% \end{equation}
For the medical code $c_i$ with the surface name $s_i$, we retrieve top-$k$ ($k=5$ in this work) passages $\cT_i$ from the corpus $\cM$ as 
\begin{equation}
\setlength{\abovedisplayskip}{8pt}
\setlength{\belowdisplayskip}{8pt}
\cT_i=\underset{d\in \cM}{\operatorname{Top-}k} \ f(s_i, d).
\label{eq:knn}
\end{equation}
The top retrieved passages are considered as the external knowledge for the medical code $c_i$.  
% our retriever's objective is to transform each of these passages into a compact, dense vector representation. This transformation allows for effective and efficient retrieval of the top-k documents that are most pertinent to the input text. 

\noindent \textbf{Summarized Knowledge Generation}. 
Although $\cT_i$ contains the most relevant information for $c_i$ from  $\cM$, directly using them to assist predictions can be suboptimal, as simply concatenating these passages often leads to long contexts, and some of the retrieved passages can also be irrelevant~\citep{yu2023chain}. 
Motivated by the fact that LLMs have strong capabilities in text summarization~\citep{zhang2023benchmarking}, we propose to use the off-the-shelf LLM (\texttt{gpt-3.5-turbo-0613}) to generate the summarized knowledge $e_i$ for medical code $c_i$ as 
\begin{equation}
\setlength{\abovedisplayskip}{8pt}
\setlength{\belowdisplayskip}{8pt}
e_i = \operatorname{LLM}([\text{Prompt}, t_{i,1}, \cdots, t_{i,k}]),
\label{eq:summ}
\end{equation}
where $t_{i}\in \cT_i$ stands for the retrieved passages in Eq.(\ref{eq:knn}). We incorporate information related to the downstream task within our prompt to ensure the generated summaries are task-specific. 
% {Please refer to Appendix~\ref{apd:prompt} for detailed prompt design.}
Detailed prompt designs can be found in Appendix~\ref{apd:prompt}.
% The prompt for this summarization step is deferred to Appendix \yy{}.

\noindent \textbf{Remark}. 
The retrieval step is efficient as the corpus indexing only needs to be done \emph{once} before applying to prediction tasks. 
It only needs one extra ANN retrieval operation per query, which is efficiently supported by FAISS~\cite{faiss}.
Besides, we cache the summarized knowledge for each medical code to avoid
redundant operations.

% \vspace{-0.7ex}
\begin{algorithm}[t]
\small
    \caption{Overview of \ours{}.}\label{alg:method}
    \begin{algorithmic}[1] % The number tells where the line numbering should start
    \STATE \textbf{Input:} $\mathcal{P}$: patients; $V$: corresponding hospital visits of patients.
        \STATE Initializing multi-source external knowledge $\mathcal{M}$;
        \FOR{$i=1,\cdots,|V|$}
            % \STATE Get the corresponding hospital visits $V$ of patient $p$;
            \FOR{$c_i\in v_i$}
                \STATE Get the medical code $c_i$ and the corresponding textual name $s_i$ included in visit $v_i$;\\
                \STATE \blue{// \textit{Passage Retrieval}}\\
                \STATE Retrieve passages $\cT_i$ via Eq.~(\ref{eq:knn})\\
                \STATE \blue{// \textit{Knowledge Summarization (Accelerated with caching)}}\\
                \STATE Summarize knowledge $e_i$ for $c_i$ via Eq.~(\ref{eq:summ}); \\
            \ENDFOR\\
            \STATE \blue{// \textit{Co-training}}\\
            \STATE Predict $\hat{y}_{i,1}$ with knowledge-augmented model $g_\phi$ via Eq.~(\ref{eq:aug});\\
            \STATE Predict $\hat{y}_{i,2}$ with visit-based local model $f_\theta$ via Eq.~(\ref{eq:visit});\\
            \STATE \blue{// \textit{Update Model Parameters}}\\
        	\STATE Compute loss function $\mathcal{L}$ via Eq.~(\ref{eq:loss2});\\
            \STATE Update model parameters $\phi$ and $\theta$;\\
        \ENDFOR\\
    \textbf{Output:} Augmented model $g_\phi$ and local model $f_\theta$; Final prediction $\tilde{y}_j=\beta\hat{y}_{j,1}+(1-\beta)\hat{y}_{j,2}$ for the $j$-th test example $p_j$.
    \end{algorithmic}
    \label{alg:main}
\end{algorithm}

\subsection{Augmenting Patient Visits with Summarized Knowledge via Co-training}
% \vspace{-0.3ex}
Recall that patient visits and summarized knowledge encode complementary information for clinical prediction tasks --- visits capture \emph{co-occurrence relationships}, while summarized knowledge encodes \emph{semantic information}. To effectively aggregate these two types of information, we design a co-training approach, detailed as follows.

\noindent \textbf{Augmented Model $g_{\phi}$  with Summarized Knowledge}.
For patient $p_i$ having the hospital visit $v_i$ with involved medical codes $\cC_i$, we decompose $\cC_i$ into three subsets: $\cC_i^{\operatorname{d}}$ for diseases, $\cC_i^{\operatorname{m}}$ for medications, and $\cC_i^{\operatorname{p}}$ for procedures.
For each type of medical code, we flatten the visit into a document by concatenating all the codes and their summarized knowledge in a reversed sequential order.
% \ran{double check}
% 
% For patient $p_i$ having the visit history $V_i$ with length $t$, we decompose $V_i$ based on three types of medical codes as $V_{i}^{\operatorname{d}}$, $V_{i}^{\operatorname{m}}$, $V_{i}^{\operatorname{p}}$, respectively. 
% Then, we flatten the visit into a document by first sorting the code in reverse chronological order, then concatenating codes and summarized knowledge together. 
For example, for disease code $\cC_i^{\operatorname{d}}$, the flattened document can be 
$X^{\operatorname{d}}_i = \left\{[\operatorname{CLS}], D_t, D_{t-1}, \ldots, D_1\right\}$, where $D_i = ||_{c\in D_i} (c, e)$ is the concatenation of disease code and its summarized knowledge (Eq.~\ref{eq:summ}) within the $i$-th visit.
We then use a pre-trained language model (PLM) with a multi-layer perceptron (MLP) classification head as $g_{\phi}$ for prediction with flattened documents as inputs:
\begin{equation}
\setlength{\abovedisplayskip}{8pt}
\setlength{\belowdisplayskip}{8pt}
\bh_{i}^{k} =\operatorname{PLM}(X_i^{k}), ~~
\hat{y}_{i,1} = \operatorname{MLP}\left( ||_{k\in\cS}\bh_{{i}}^{{k}}\right). 
% \\ 
% \bh_{{i}}^{\operatorname{m}}, \bh_{{i}}^{\operatorname{p}}]).
\label{eq:aug}
\end{equation}
Here $\cS=\{\operatorname{p},\operatorname{m},\operatorname{d}\}$, $\bh_i$ is the representation of \texttt{[CLS]} token of $X_i$, $\hat{y}_{i,1}$ is the prediction for the target task.
We share PLM weights for three types of medical codes to improve efficiency.

% 2.2.1 Model Inputs. As introduced in Section 2.1, for each item $i$ and corresponding attribute dictionary $D_i$, we flatten the dictionary into an item "sentence" $T_i=\left\{k 1, v 1, k 2, v 2, \ldots, k_m, v_m\right\}$ where $k$ and $v$ are described by words, formally $(k, v)=\left\{w_1^k, \ldots, w_c^k, w_1^v, \ldots, w_c^v\right\}$. To encode a user's interaction sequence $s=\left\{i_1, i_2, \ldots, i_n\right\}$, we first reverse items in a sequence to $\left\{i_n, i_{n-1}, \ldots, i_1\right\}$ because intuitively recent items (i.e., $i_n, i_{n-1}, \ldots$ ) are important for the next item prediction and reversed sequences can make sure recent items are included in the input data. Then, we use the item "sentences" to replace items and add a special token [CLS] at the beginning of sequences. Hence, model inputs are denoted as:
% $$
% X=\left\{[\mathrm{CLS}], T_n, T_{n-1}, \ldots, T_1\right\}
% $$
% where $X$ is a sequence of words containing all items and corresponding attributes the user interacted with in the historical interactions.
% we flatten the dictionary into an item “sentence”

\noindent \textbf{Local Model $f_{\theta}$ with Visit Information}. To harness the visit-level information, various deep learning architectures have been proposed. 
In principle, $g_{\phi}$ can be combined with any $f_{\theta}$ to improve performance. In main experiments, we use a hypergraph transformer (HyGT,~\citet{pmlr-v193-xu22a,cai2022hypergraph}) due to its strong ability to capture high-order relationships between visits and medical codes. It first builds hypergraphs $\cG=(\cV,\cE)$ by treating medical codes as nodes and patients as hyperedges, then leverages self-attention for aggregating neighborhood information. The details for HyGT are in Appendix~\ref{apd:hygt}. We obtain the prediction $\hat{y}_{i,2}$ with $f_\theta$ as 
\begin{equation}
\setlength{\abovedisplayskip}{8pt}
\setlength{\belowdisplayskip}{8pt}
\be_{i} =\operatorname{HyGT}(\cG, V_i), 
\hat{y}_{i,2} = \operatorname{MLP}(\be_i),
\label{eq:visit}
\end{equation}
where $\be_{i}$ is the representation of patient $i$ after hypergraph transformer.

\noindent \textbf{Co-training}. 
We integrate the two predictors into a co-training framework, with the learning objective:
\begin{equation}
\begin{aligned}
\setlength{\abovedisplayskip}{8pt}
\setlength{\belowdisplayskip}{8pt}
\cL_{\text{aug}} = &\mathbb{E}_{(V_i, y_i) \sim \cP}~\ell(\hat{y}_{i,1}, y_i) + \lambda \cD_{\operatorname{KL}}(\hat{y}_{i,1}, \tilde{y}), \\
% \ell(\hat{y}_{i,2}, y_i) \\ 
\cL_{\text{loc}} =&\mathbb{E}_{(V_i, y_i) \sim \cP}~\ell(\hat{y}_{i,2}, y_i) + \lambda \cD_{\operatorname{KL}}(\hat{y}_{i,2}, \tilde{y}),
\end{aligned}\label{eq:loss2}
\end{equation}
where $\ell(\cdot)$ is the binary cross-entropy loss, $\tilde{y} = \beta\hat{y}_{i,1} + (1-\beta)\hat{y}_{i,2}$, $\lambda, \beta$ are two hyperparameters. Two losses in Eq.~\ref{eq:loss2} are designed to encourage $f_\theta$ and $g_\phi$ regularize each other, which can stabilize the learning for two models.
During the \textbf{inference stage}, we directly use the $\tilde{y}_j$ as the final prediction for the $j$-th test example $p_j$. 
The overall algorithm of \ours{} is illustrated in Algorithm~\ref{alg:main}.
\section{Experiments}
\label{sec:exp}
\subsection{Experiment Setups}
\textbf{Datasets.}
\label{apd:task}
We present the detailed statistics of MIMIC-III and \dataset{} in Table~\ref{tab:data_stats}.

\paragraph{MIMIC-III.}
The MIMIC-III dataset~\citep{johnson2016mimic} 
is a large, freely available database that contains de-identified health-related data from over 4,000 patients who stayed in critical care units at the Beth Israel Deaconess Medical Center between 2001 and 2012. 
We conduct the phenotyping prediction task proposed by~\cite{harutyunyan2019multitask}. It aims to predict whether the 25 pre-defined acute care conditions (see Table~\ref{tab:mimic_phenotypes}) are present in a patient's next visit, based on the information from their current visit. The problem is formulated as a 25-label binary classification, considering that multiple phenotypes may exist in a single visit.
For data preprocessing, we focus on patients with multiple hospital visits, identified based on their admission information. We extract pairs of consecutive visits for each patient. For each pair, we extract diseases, medications, and procedures from the health records in the former visit as input, and identify the phenotypes in the latter visit as labels, using Clinical Classifications Software
(CCS) from the Healthcare Cost and Utilization Project (HCUP)\footnote{\url{https://hcup-us.ahrq.gov/toolssoftware/ccs/AppendixASingleDX.txt}}.

\paragraph{\dataset.}
For the {\dataset} dataset, we conduct a CVD outcome prediction task, which predicts whether patients with type 2 diabetes will experience CVD complications within 1 year after their initial diagnosis, including coronary heart disease (CHD), congestive heart failure (CHF), myocardial infarction (MI), or stroke. Diseases are identified by their ICD-9 or ICD-10 clinical codes.
The usage of data has been approved by the Institutional Review Board (IRB).

\begin{table}[t]
\caption{The statistics of MIMIC-III and {\dataset}.}
\label{tab:data_stats}
\centering
\renewcommand\arraystretch{0.9}
\resizebox{0.8\linewidth}{!}{
\begin{tabular}{lcc}
\toprule
\bfseries Stats & \bfseries MIMIC-III    & \bfseries \dataset          \\ 
\midrule 
\# of diagnosis & 846 & 7915\\
\# of medication & 4525 & 489\\
\# of procedure & 2032 & 4321\\
\# of health records & 12353 & 36611 \\
\bottomrule
\end{tabular}
}
\end{table}
% \textbf{Datasets.}
% We conduct experiments on the public MIMIC-III dataset~\citep{johnson2016mimic} and a private {\dataset} dataset collected from a large healthcare system in the United States.
% We perform a 25-label phenotypes prediction task on MIMIC-III, and a cardiovascular disease (CVD) endpoints prediction task for diabetes patients on {\dataset}. We randomly split them into train/validation/test sets by 7:1:2. Please refer to Appendix~\ref{apd:task} for details. 

% Uncomment and define \dataset and \ours or replace them with actual content
% \newcommand{\dataset}{Your Dataset Name Here}
% \newcommand{\ours}{Your Model Name Here}

\begin{table*}[t]
% \setlength{\tabcolsep}{1pt}
% \resizebox{\linewidth}{!}{
% \floatconts
\vspace{-1ex}
\caption{Performance on two EHR datasets compared with baselines. The result is averaged over five runs. We use * to indicate statistically significant results ($p<0.05$).
For `w/o Retrieval', we directly use LLM to generate summarized knowledge. For `w/o LLM Summarization', we concatenate top-$k$ retrieved documents as summarized knowledge.
`w/ $g_\phi$ only' means we set $\lambda=0, \beta=1$ (i.e., only use the prediction from $g_\phi$ as the final prediction).
\vspace{-1.5ex}
}
\label{tab:main_results}
%   \begin{adjustbox}{width=\columnwidth,center}
%   \resizebox{\columnwidth}{!}{
% \vspace{-12pt}\
\centering
\renewcommand\arraystretch{0.92}
\resizebox{0.99\linewidth}{!}{
\begin{tabular}{lcccccccc}
\toprule
\multirow{2.5}{*}{\bfseries Model} & \multicolumn{4}{c}{\bfseries MIMIC-III}    &  \multicolumn{4}{c}{\bfseries \dataset}           \\ 
\cmidrule(lr){2-5} \cmidrule(lr){6-9}
&  \bf ACC &   \bf AUROC &  \bf AUPR &  \bf F1 & 
  \bf ACC &  \bf AUROC &  \bf AUPR &  \bf F1\\
\midrule
% LR~\citep{menard2002applied} & 68.66 & 64.62 & 45.63 & 13.74 & 76.22 & 57.22 & 25.99 & 42.18\\
% SVM~\citep{cortes1995support} & 72.02 & 55.10 & 34.19 & 32.35 & 68.57 & 53.57 & 23.50 & 52.34\\
Transformer~\citep{li2020behrt} & 76.18 & 80.61 & 67.12 &42.75  & 78.10 & 69.49 & 40.14 & 58.23 \\ 
GCT~\citep{choi2020learning} &
77.20 & 78.62 & 64.87 & 37.57 & 76.51 &	68.31 &	37.55 &	44.10 \\
HyGT~\citep{xu2023hypergraph} & {78.07} & 81.09 & 68.08 & \underline{44.93} & \underline{79.45} & 70.59 & 41.04 & 60.00\\ %\hline
% GAT~\citep{gat} & 77.47 & 79.89 & 67.14 & 34.88 & 78.02 & 67.50 & 36.06 & 57.39\\ \midrule
% Dr.Knows~\citep{gao2023leveraging} & 77.93 & 80.63 & 68.38 & \underline{40.04} & 76.77 & 68.29 & 37.93 & 58.05\\
% MedPath~\citep{ye2021medpath} & 75.31 & 78.14 & 67.62 & 38.94 &  --- & --- & --- & ---\\
MedRetriever~\citep{ye2021medretriever} & 77.15 & 80.14 & 68.45 & 39.29 & 78.95 & 70.07 &  42.19 & 57.96 \\
SeqCare~\citep{xu2023seqcare} & 77.44 & 80.98 & 68.53 & 39.16 & --- & --- & --- & ---\\
GraphCare~\citep{jiang2023graphcare} & \underline{80.11} & {82.26} & {71.19} & 44.33 & 79.09 & \underline{71.12} & \underline{43.98} & 59.00 \\%\hline
CORE~\citep{van-aken-etal-2021-clinical} & 79.63 & 82.05 & 70.79 & 43.76 & 77.11 & 67.84 &   40.74 & \underline{61.12} \\
BEEP~\citep{naik-etal-2022-literature} & 79.90 & \underline{82.67} & \underline{71.58} & 44.15 & 79.29 & 68.59 & 41.93 & 60.95 \\\midrule
% 78.98 & \underline{82.56} & \underline{71.09} & 39.33 & 79.49 & 69.60 & \underline{41.48} & 56.65 
\rowcolor{magenta!10} {\ours} & \textbf{81.59}* (1.8\%) &	\textbf{84.97}*  (2.8\%) &	\textbf{74.64}*  (4.3\%) &	\textbf{48.19}* (7.2\%) & 
\textbf{80.41}*  (1.2\%) & \textbf{73.80}*  (3.8\%) & \textbf{48.40}*  (10.1\%) & \textbf{63.98}*  (4.7\%)\\
~~ w/o Retrieval  & 80.68 & 83.29 & 72.95 & 44.65 & 79.83 & 73.06 & 47.05 & 63.25 \\
~~ w/o LLM Summarization & 80.08 & 82.14 & 71.35 &41.49 &77.30 & 69.71 &42.58 & 61.70 \\
~~ w/ Augmented Model $g_{\phi}$ Only & 81.04 &	83.80	& 73.41 &	46.83 & 79.70	& 73.15	& 47.62	 & 63.33\\
\bottomrule
\end{tabular}
}
% \vspace{-1.4ex}
% }
% \end{adjustbox}
\end{table*}

% \clearpage

\noindent \textbf{Evaluation Metrics.}
Following \citet{choi2020learning}, we employ Accuracy, AUROC, AUPR, and Macro-F1 as evaluation metrics, where AUROC is the main metric.
For accuracy and F1 score, we use a threshold of 0.5 after obtaining predicted results.

\noindent \textbf{Baselines.} 
We consider the following baselines, with a focus on knowledge-enhanced EHR predictive models:
\begin{itemize}
    \item \textbf{Transformer}~\citep{li2020behrt}: It leverages the Transformer~\citep{NIPS2017_3f5ee243} architecture to model sequential EHR visits for clinical prediction tasks.
    \item \textbf{GCT}~\citep{choi2020learning}: It employs the Transformer model to learn the EHR's hidden structure via medical codes. Additionally, it introduces the Graph Convolutional Transformer, integrating graph neural networks to utilize the EHR structure for prediction. 
    \item \textbf{HyGT}~\citep{xu2023hypergraph}: It leverages hypergraph transformers that regard patients as hyperedges and medical codes as nodes for EHR predictive tasks.
    \item \textbf{MedRetriever}~\citep{ye2021medretriever}: It retrieves the most relevant text segments from a local medical corpus using string similarity. Then, it uses query features aggregated with EHR embeddings and disease-specific documents via self-attention for disease prediction tasks. 
    \item \textbf{SeqCare}~\citep{xu2023seqcare}: It is one of the strongest models for knowledge-graph-based EHR prediction with pruned graph structure learning and debiased training techniques.
    \item \textbf{GraphCare}~\citep{jiang2023graphcare}: It generates personalized knowledge graphs via prompting LLMs and leverages attention-based graph neural networks for healthcare predictions.
    \item \textbf{CORE}~\citep{van-aken-etal-2021-clinical}: It integrates clinical knowledge with specialized outcome pre-training, and uses language models to predict clinical notes for prediction.
    \item \textbf{BEEP}~\citep{naik-etal-2022-literature}: It augments the language models with the retrieved PubMed articles and fuses them with information from notes to predict clinical outcomes.
\end{itemize}

 \begin{figure*}[t]
	\centering
	\subfigure[Effect of $g_\phi$ on MIMIC-III]{
		\includegraphics[width=0.23\linewidth]{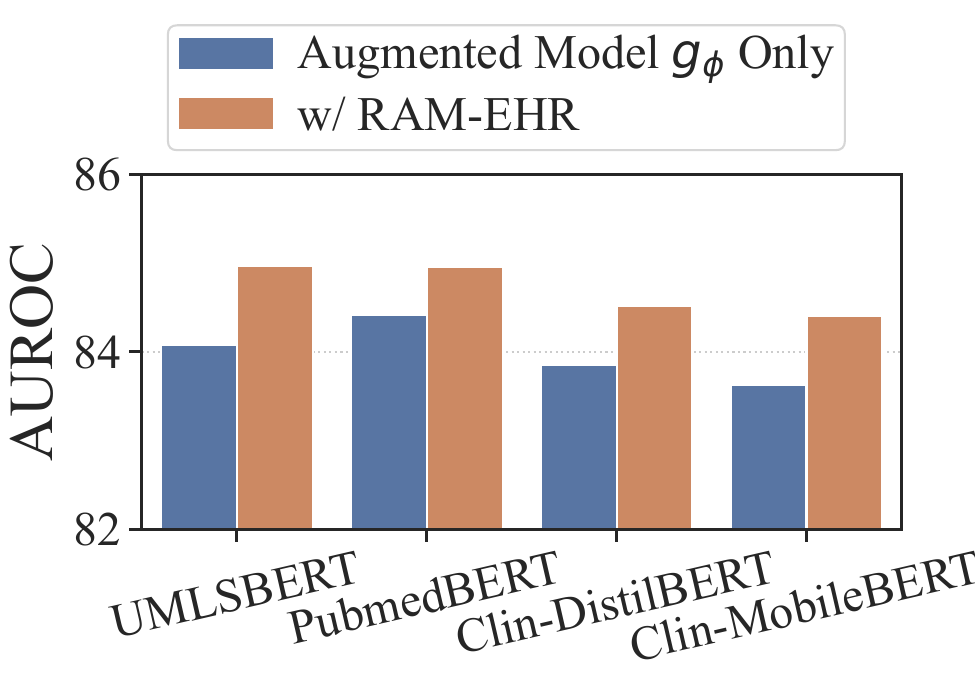}
		\label{fig:diff_classifier_mimic}
	} %\hfill
         \hspace{-0.5ex}
        \subfigure[Effect of $g_\phi$ on \dataset]{
		\includegraphics[width=0.23\linewidth]{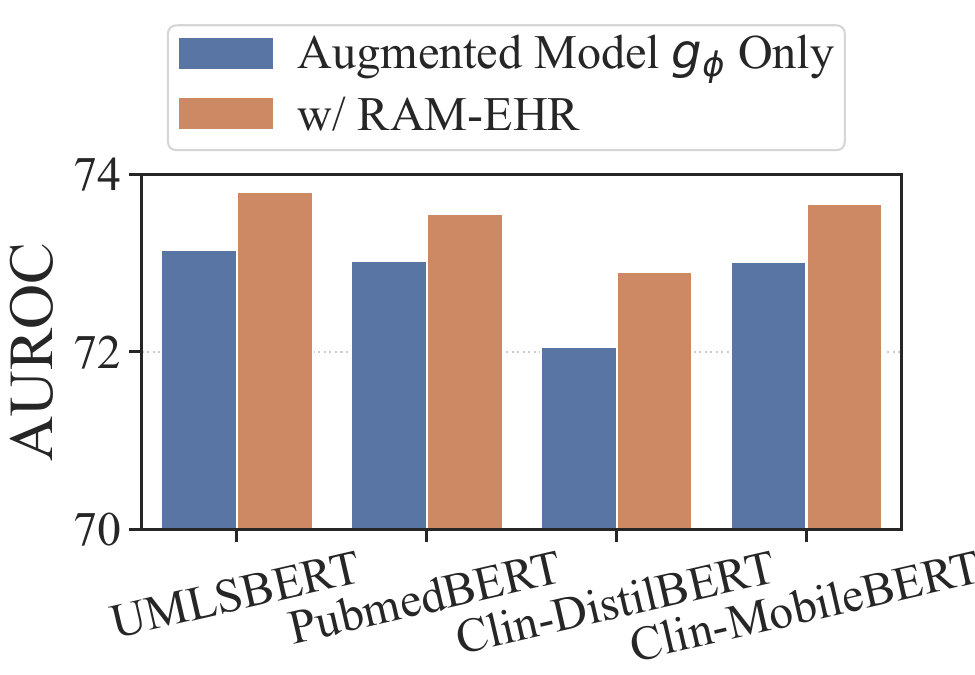}
		\label{fig:diff_classifier_cradle}
	} %\hfill
         \hspace{-0.5ex}
        \subfigure[Effect of $f_\theta$ on MIMIC-III]{
		\includegraphics[width=0.23\linewidth]{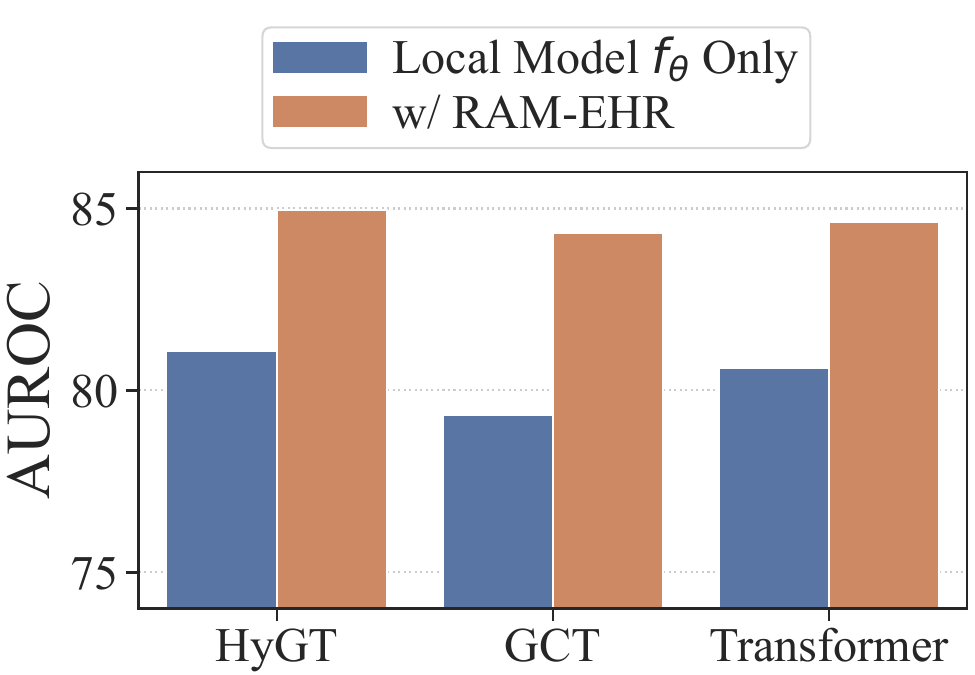}
		\label{fig:diff_gnn_mimic}
	}
          \hspace{-0.5ex}
        \subfigure[Effect of $f_\theta$ on \dataset]{
		\includegraphics[width=0.23\linewidth]{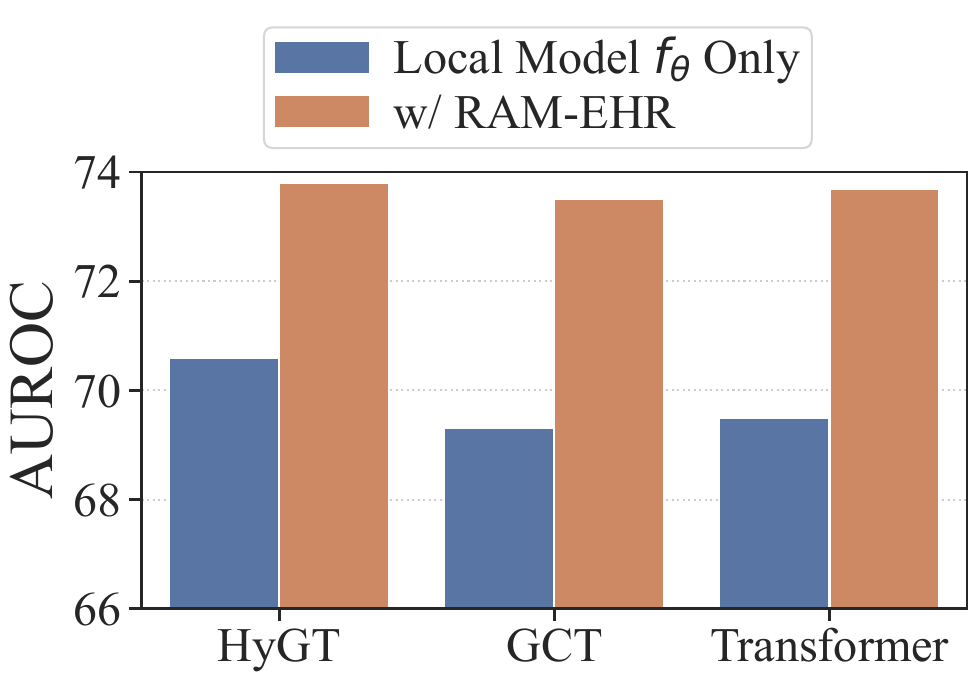}
		\label{fig:diff_gnn_cradle}
	}
        \vspace{-1ex}
	\caption{Effect of $g_\phi$ and $f_\theta$ on both datasets. \vspace{-1.5ex}}
\label{fig:effect_f_g}
\end{figure*}

\noindent \textbf{Implementation Details.} 
In this work, we use Dragon~\citep{lin-etal-2023-train} as the retriever for both passage $R_{D}(\cdot)$ and queries  $R_Q(\cdot)$\footnote{\url{https://huggingface.co/facebook/dragon-plus-{query,context}-encoder}}. We set $k=5$ during the retrieval stage.  
We choose UMLS-BERT~\citep{michalopoulos-etal-2021-umlsbert}  as the augmented model $g_{\phi}$ with 110M parameters for \ours{} and relevant baselines as $g_{\phi}$, with a maximum length of 512, and HyGT~\citep{xu2023hypergraph} as $f_{\theta}$ in main experiments. 
We also evaluate the effects of $g_{\phi}$ on PubmedBERT~\citep{Biomedbert}, Clin-MobileBERT, and Clin-DistilBERT~\citep{mobilebert}, as well as the effects of  $f_\theta$ on GCT~\citep{choi2020learning} and Transformer~\citep{li2020behrt} in Section~\ref{sec:add}.
We set the learning rate to 5e-5 for $g_{\phi}$ and 1e-4 for $f_{\theta}$, batch size to 32, and the number of epochs to 5. 
We select $\beta, \lambda$ based on the performance of the validation set. 
All the experiments are conducted on a Linux server with one NVIDIA A100 GPU.

\subsection{Main Experimental Results}

Table \ref{tab:main_results} exhibits the experiment results of \ours{} and baselines. 
\textbf{First}, we observe \ours{} 
surpasses baselines lacking external knowledge, highlighting the benefits of retrieval augmentation. 
\textbf{Second}, \ours{} outperforms knowledge-enhanced baselines due to the diverse collection of external knowledge as well as the co-training scheme 
that leverages information from both visit and semantic perspectives.
% to effectively harness the complementary information from visit and semantic views.
\textbf{Third}, directly using medical notes leads to inferior outcomes due to potential irrelevance, whereas combining medical codes with summarized knowledge as \ours{} proves more effective for prediction tasks.

% \vspace{-0.7ex}
\subsection{Additional Studies}
% \vspace{-0.4ex}
\label{sec:add}
% 1. main
\subsubsection{Ablation Study} 
On the bottom of Table~\ref{tab:main_results}, we inspect different components in \ours{} and observe that removing any of them hurts the performance, which justifies the necessity of our designs. 
As an illustration, eliminating the retrieval module might result in less informative summarized knowledge. Conversely, excluding the LLM summarization module (i.e., relying solely on retrieved text) could yield lengthy outputs and fail to incorporate all the condensed knowledge about medical codes into the augmented model. 
Besides, we observe that using the summarized knowledge with $g_\phi$ already achieves strong performance, 
highlighting the benefit of capturing the semantics of medical codes.

\begin{figure}[t]
	\centering
        \vspace{-1ex}
 	\subfigure[Effect of $\cM$, MIMIC-III]{
		\includegraphics[width=0.46\linewidth]{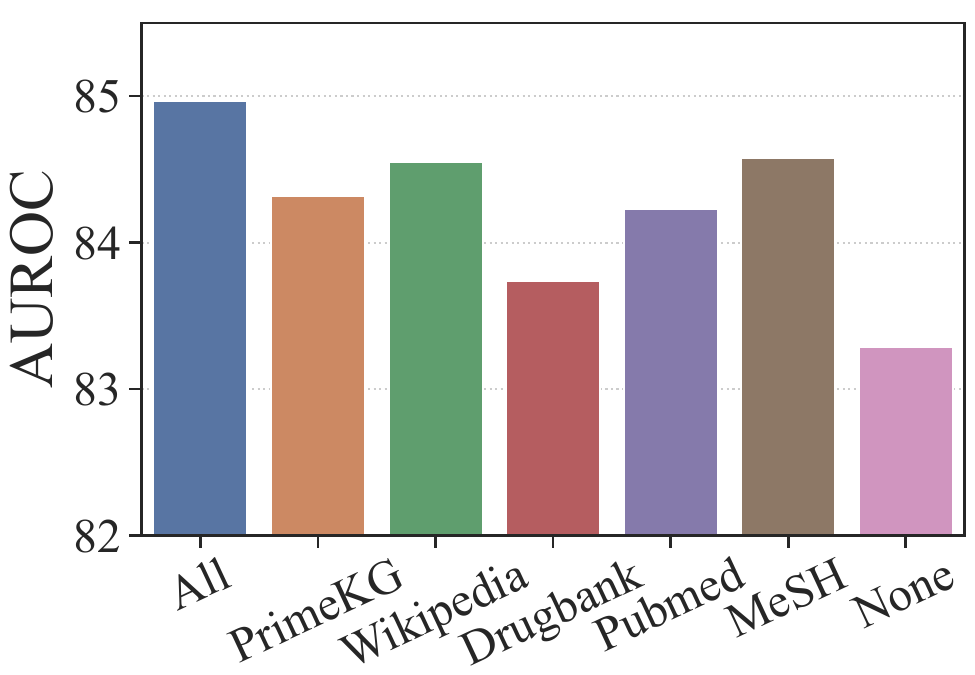}
		\label{fig:diff_knowledge_mimic}
	} %\hfill
         \hspace{-0.5ex}
        \subfigure[Effect of $\cM$, \dataset]{
		\includegraphics[width=0.46\linewidth]{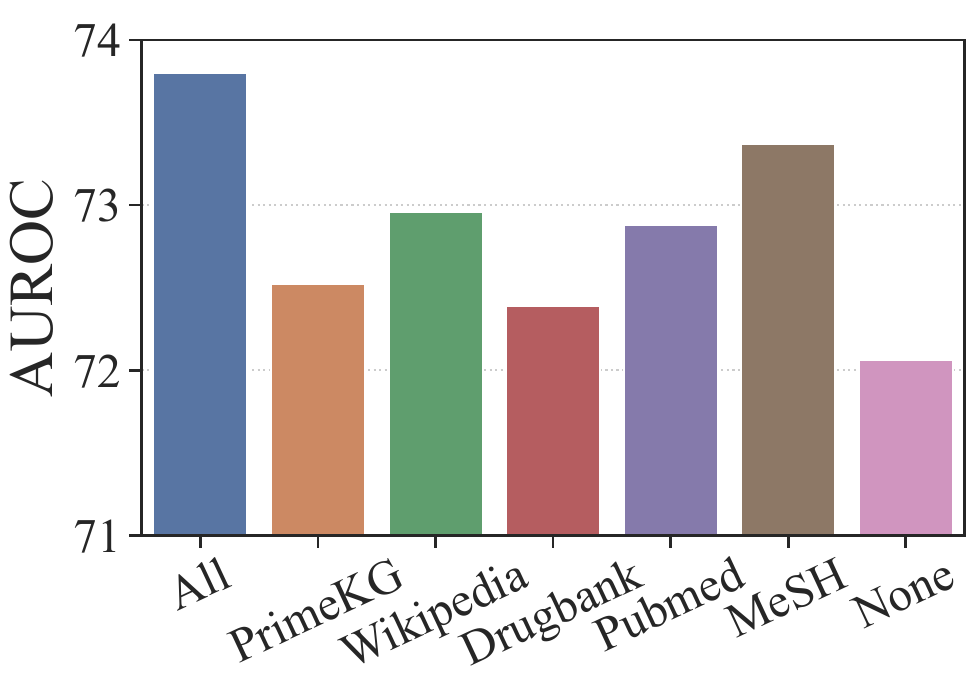}
		\label{fig:diff_knowledge_cradle}
	} %\hfill
        \vspace{-1ex}
	\caption{Studies on Information Source $\cM$.\vspace{-1.5ex}}
\label{fig:effect_M}
\end{figure}

\noindent \subsubsection{Effect of $f_\theta$ and $g_\phi$}
We evaluate the effect of the augmented model $g_\phi$ and the local model $f_\theta$ on both datasets in Figure~\ref{fig:effect_f_g}. The experimental results demonstrate that both models contribute to the performance gain. Moreover, it is observed that {\ours} is flexible to be applied upon different models, with a comparable performance.
% Figure~\ref{fig:diff_classifier_mimic}, \ref{fig:diff_gnn_mimic} demonstrate \ours{}'s performance with various $f_\theta$ and $g_\phi$, showing consistent improvements across models, which confirms \ours{}'s flexibility.
% With various $f_\theta$ and $g_\phi$, we demonstrate the flexibility of \ours{} in Figure~\ref{fig:diff_classifier_mimic} and \ref{fig:diff_gnn_mimic} by the consistent performance gain across different models.
% Figure~\ref{} illustrates the performance of \ours{} with different $f_\theta$ and $g_\phi$ as backbone. 
% We notice that the gain is consistent across different models, indicating the flexibility of \ours{}. 
% With a lightweight $f_\theta$ using only 25M parameters (Clin-MobileBERT), \ours{} still achieves a competitive performance (99.6\% of full performance), offering an efficient solution for EHR predictive models.
Notably, even with a lightweight $f_\theta$ (Clin-MobileBERT) having only 25M parameters, \ours{} reaches close performance to UMLS-BERT, providing an efficient option for EHR predictive modeling.

\noindent \subsubsection{Effect of Information Source $\cM$} 
% To assess the effectiveness of various components within $\cM$, we evaluate the performance of each knowledge source. 
We then evaluate the effectiveness of each knowledge source within $\cM$.
Figure~\ref{fig:effect_M} indicates that incorporating all corpus yields the highest performance, highlighting the value of diverse corpora. Besides, using Drugbank alone contributes minimally, likely due to its limited scope of medication information. Moreover, we observe that leveraging knowledge bases (e.g., MeSH) is more beneficial than literature sources, as they offer broader and more generic information for EHR prediction tasks.
% To examine the utility of different components in $\cM$, we exhibit the performance using each knowledge source in $\cM$. 
% From the result, we observe that including all resources archives best performance, justifying the complementary information from different corpora. Besides, using Drugbank only leads to smallest gains, as it only contains medication information and is less comprehensive. Besides, we observe that using knowledge bases are more helpful than literatures, as it contains more generic knowledge that are helpful for clinical prediction tasks.

\begin{figure}[t]
	\centering
        \vspace{-1ex}
	\subfigure[$\beta$ in Eq.~\ref{eq:loss2}]{
		\includegraphics[width=0.46\linewidth]{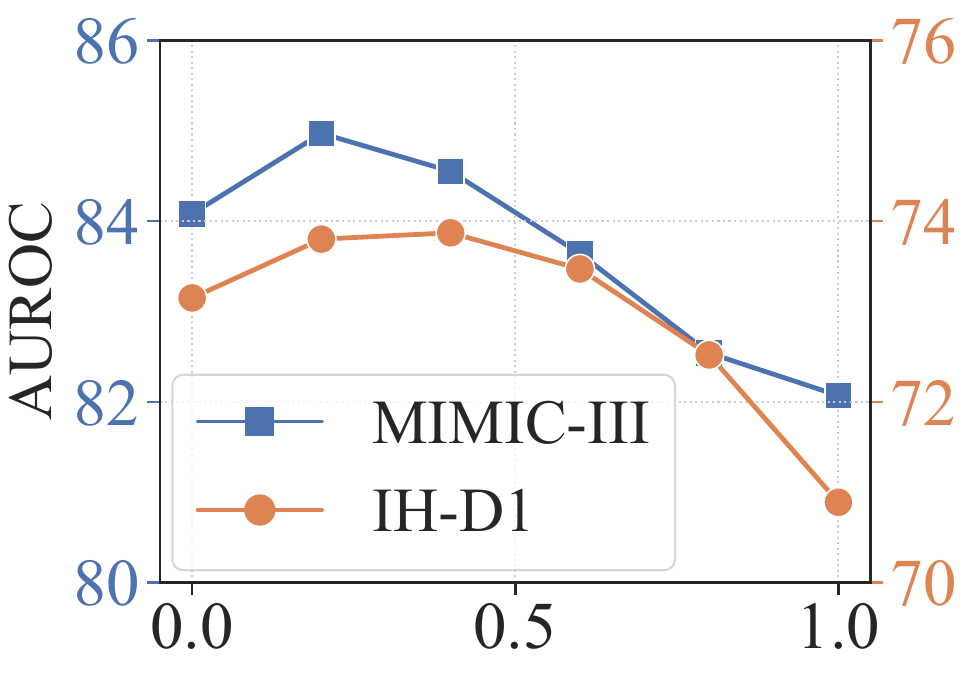}
		\label{fig:beta}
	} %\hfill
         \hspace{-0.5ex}
        \subfigure[$\lambda$ in Eq.~\ref{eq:loss2}]{
		\includegraphics[width=0.46\linewidth]{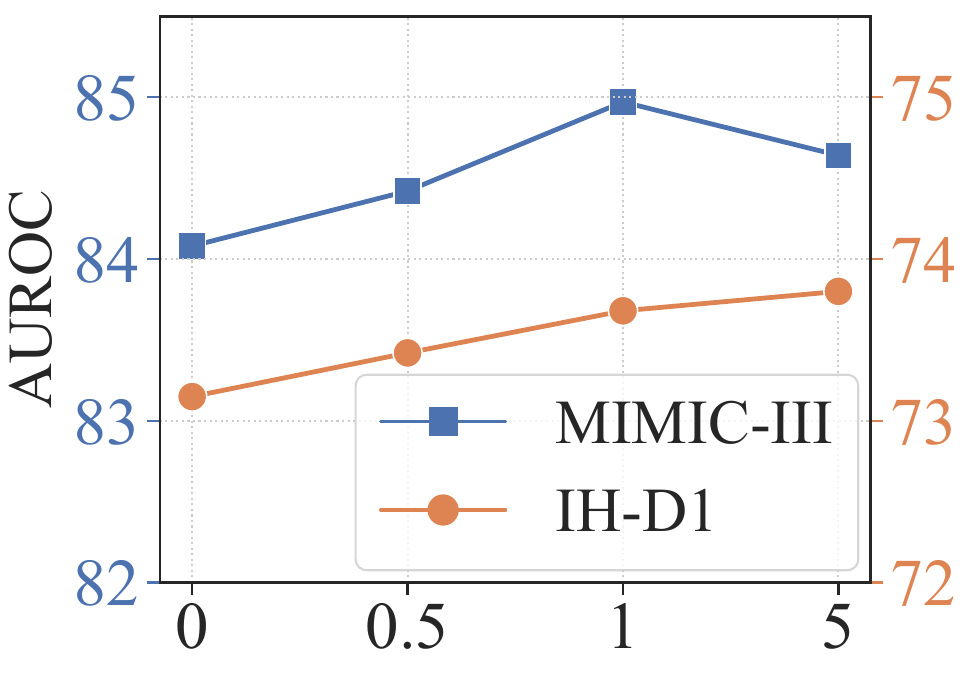}
		\label{fig:lambda}
	}
        \vspace{-1ex}
	\caption{Parameter studies of $\beta$ and $\lambda$ on both datasets.\vspace{-1.5ex}}
\label{fig:parameter}
\end{figure}

\begin{figure*}[ht]
	\centering
	\subfigure[Case Study]{
		\includegraphics[width=0.61\linewidth]{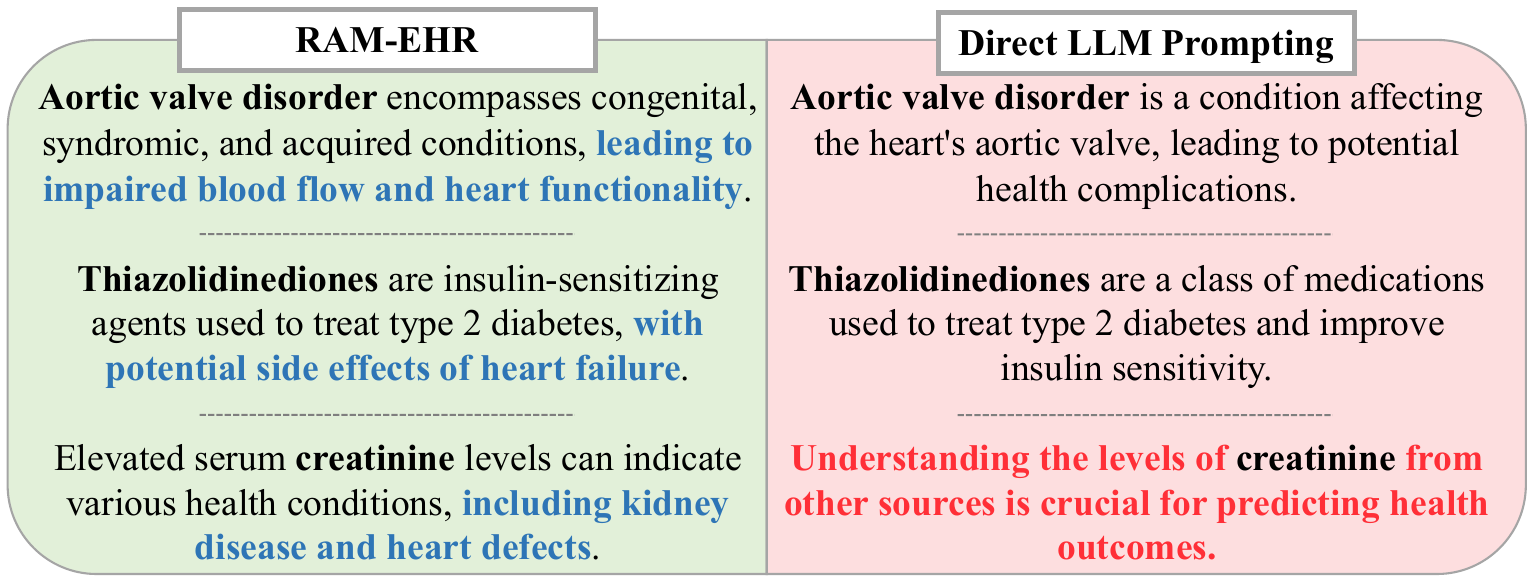}
		\label{fig:case_study}
	} %\hfill
         \hspace{-0.5ex}
        \subfigure[Human Study]{
		\includegraphics[width=0.33\linewidth]{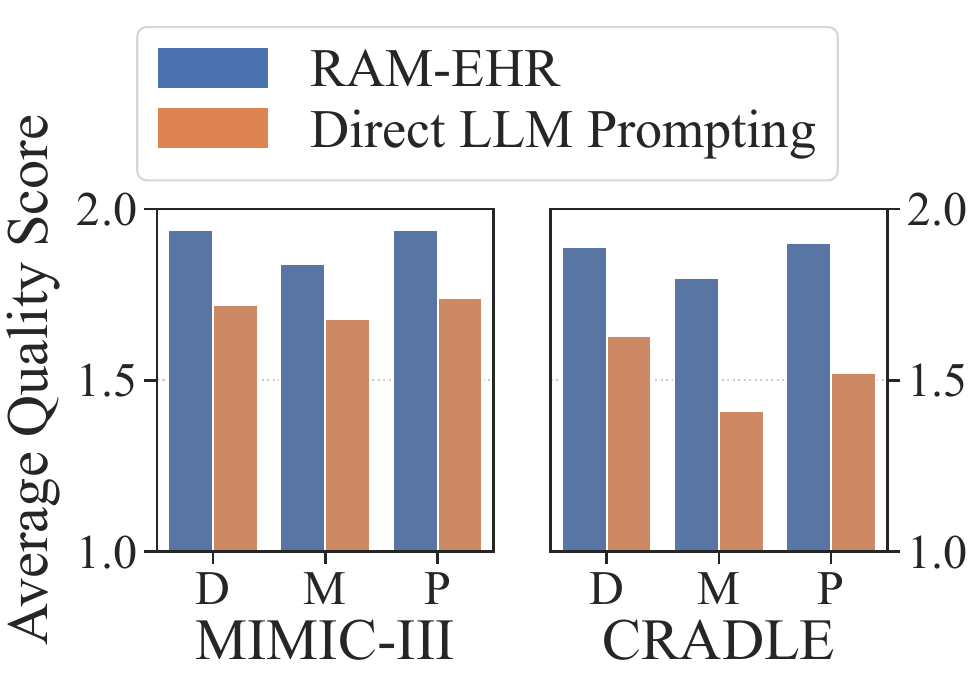}
		\label{fig:case_quality}
	}
        \vspace{-1ex}
	\caption{Case study and human study. The case study compares knowledge summarized by {\ours} and directly generated by LLM prompting. \textbf{Bold} denotes disease, medication and procedure concepts.   \textcolor{blue}{\textbf{Blue}} and \textcolor{red}{\textbf{Red}} indicate useful and irrelevant knowledge.}
  \vspace{-2ex}
\label{fig:case_study}
\end{figure*}

\noindent \subsubsection{Parameter Study}
\label{sec:para}
% We evaluate the effect of the augmented model $g_\phi$, the local model $f_\theta$, and the knowledge source $\cM$ on \dataset{} in Figure~\ref{fig:ablation_cradle}. The experimental results further demonstrate that both models and different knowledge sources contribute to the performance gain. Moreover, it is observed that {\ours} is flexible to be applied upon different models, with a comparable performance with {\ours}.

In Figure~\ref{fig:parameter}, we conduct parameter studies on both datasets for $\beta$ and $\lambda$ in Eq.~\ref{eq:loss2}.
Figure~\ref{fig:beta} demonstrates that the model achieves the best performance when $\beta$ is set to 0.2 and 0.4 on MIMIC-III and \dataset, respectively, while the gain diminishes at the extremes. This highlights the contribution of combining the predictions from both the augmented model and the local model on the performance gain. In addition, $\lambda$ is set to 1 and 5 on MIMIC-III and \dataset, respectively, according to Figure~\ref{fig:lambda}. The positive values of $\lambda$ indicate that the consistency loss enhances model performance.

\noindent \subsection{Case Study}
Figure~\ref{fig:case_study} presents a case study on {\dataset} to compare knowledge summarized by {\ours} and directly generated by LLM prompting. We observe that {\ours} provides more relevant information for the downstream task, particularly regarding the CVD outcome in this case, compared to direct LLM prompting. 
This also aligns with the \emph{human study} evaluating the quality of 40 randomly sampled knowledge per type of code on a scale of [0,1,2] in Figure~\ref{fig:case_quality}. 
Specifically, a score of 0 represents irrelevant knowledge, 1 denotes somewhat useful knowledge and 2 indicates highly useful knowledge. The detailed guidelines provided for the annotators are presented in Appendix~\ref{apd:additional}.

% \vspace{-0.8ex}
\section{Conclusion}
% \vspace{-0.6ex}
% \ours{} introduces a retrieval augmentation pipeline to boost clinical predictions in Electronic Health Records (EHRs) by utilizing multiple knowledge sources and dense retrieval for information related to medical concepts. This approach, combined with consistency regularization, enhances EHR predictive models by integrating patient visit data and summarized knowledge. Experiments demonstrate \ours{}'s superiority over existing methods, achieving notable improvements in prediction accuracy.
We propose \ours{}, which uses dense retrieval with multiple knowledge sources and consistency regularization to enhance EHR prediction tasks. 
% Experimental results on two EHR datasets demonstrate the superiority of RAM-EHR over baseline methods,
% In this work, we propose \ours{}, a retrieval-augmented framework for EHR prediction tasks. \ours{} leverages multiple knowledge sources and dense retrieval for medical concepts, integrated with consistency regularization.
Experiments on two EHR datasets show the efficacy of \ours{} over baselines with a gain of 3.4\% in  AUROC and 7.2\% in AUPR.
Human studies confirm the usefulness of generated knowledge.

\section*{Acknowledgement}
We thank the anonymous reviewers and area chairs for  valuable feedbacks. 
This research was partially supported by the internal funds and GPU servers provided by the Computer Science Department of Emory University. 
JH was supported by NSF grants IIS-1838200 and IIS-2145411.

\section*{Limitations}
In this work, we propose \ours{} to unify external knowledge in text format and adapt it for EHR predictive tasks. Despite its strong performance, we have listed some limitations of \ours{}:

\paragraph{Building Multi-source Corpus $\cM$.} 
In this study, we construct a multi-source corpus $\cM$ by manually selecting five relevant sources within the clinical domain. In real-world scenarios, the grounding corpora usually require customization according to query domains and user needs. Therefore, effectively selecting grounding corpora and efficiently evaluating their relative contributions remains an unresolved issue. Furthermore, retrieved evidence may contain noise that could potentially degrade model performance, highlighting the importance of developing fine-grained filtering or re-ranking modules as a crucial area for future research.

% These analyses will extend beyond our empirical settings and reveal a wider application scenario of MoMA.
 
\paragraph{Other Types of Models and Tasks.} 
Although we have tested three $f_{\theta}$ and four $g_{\phi}$ combinations on two clinical prediction tasks, there are numerous other clinical tasks and prediction models that exist. 
Due to computational constraints, we are unable to explore all possible model combinations. Extending \ours{} to encompass additional models and tasks is a crucial avenue for future research.
% \ran{}, other tasks
% \paragraph{Addition Models}
% Due to the limitation of computational resources, we have not evaluated t

\paragraph{Efficiency.}
The integration of the augmented model $g_\phi$ can result in additional time complexity. 
In our main experiment setups (using UMLS-BERT), co-training usually takes $1.5\times$ to $2\times$ more times than using the local model alone. One potential solution is to use a lightweight model (e.g., Clin-MobileBERT) to improve efficiency.

\section*{Ethical Considerations}
One potential ethical consideration concerns the use of credential data (MIMIC-III and \dataset) with GPT-based online services. We have signed and strictly adhered to the PhysioNet Credentialed Data Use Agreement\footnote{\url{https://physionet.org/about/licenses/physionet-credentialed-health-data-license-150/}} for the legal usage of the MIMIC-III dataset. To prevent sensitive information from being shared with third parties through APIs, we carefully follow the guidelines\footnote{\url{https://physionet.org/news/post/gpt-responsible-use}} for the responsible use of MIMIC data in online services. Specifically, we have requested to opt out of human review of the data by filling out the Azure OpenAI Additional Use Case Form\footnote{\url{https://aka.ms/oai/additionalusecase}} in order to utilize the Azure Open AI service while ensuring that Microsoft does not have access to the patient data.
The utilization of LLMs in our framework is strictly to build medical concept-specific KGs. 
In addition, the building of medical concept-specific KGs \emph{does not involve direct interaction} with any individual patient information. We iterate through all concepts in the medical coding system (e.g., CCS and ICD) to generate their respective KGs using LLMs, and these KGs are stored locally.

\bibliography{custom,anthology}
\clearpage
\appendix

\section{Phenotypes}
\begin{table}[t]
\caption{The 25 pre-defined phenotypes in MIMIC-III.}
\label{tab:mimic_phenotypes}
\centering
\renewcommand\arraystretch{0.9}
\resizebox{0.99\linewidth}{!}{
  \begin{tabular}{ll}
  \toprule
  \bfseries Phenotype & \bfseries Type \\
  \midrule
Acute and unspecifed renal failure & acute \\
Acute cerebrovascular disease & acute \\
Acute myocardial infarction & acute \\
Cardiac dysrhythmias & mixed \\
Chronic kidney disease & chronic \\
Chronic obstructive pulmonary disease & chronic \\
Complications of surgical/medical care & acute \\
Conduction disorders & mixed \\
Congestive heart failure; nonhypertensive & mixed \\
Coronary atherosclerosis and related & chronic \\
Diabetes mellitus with complications & mixed \\
Diabetes mellitus without complication & chronic \\
Disorders of lipid metabolism & chronic \\
Essential hypertension & chronic \\
Fluid and electrolyte disorders & acute \\
Gastrointestinal hemorrhage & acute \\
Hypertension with complications & chronic \\
Other liver diseases & mixed \\
Other lower respiratory disease & acute \\
Other upper respiratory disease & acute \\
Pleurisy; pneumothorax; pulmonary collapse & acute \\
Pneumonia & acute \\
Respiratory failure; insufficiency; arrest & acute \\
Septicemia (except in labor) & acute \\
Shock & acute \\
  \bottomrule
  \end{tabular}
}
\end{table}
The details of 25 phenotypes are listed in table~\ref{tab:mimic_phenotypes}.

\section{Knowledge Translating Format}
\label{apd:format}
We list the template to transform knowledge triplets into sentences in KG as follows:
\begin{lstlisting}[linewidth=\columnwidth,breaklines=true]
candidate_relation = ["disease_phenotype_positive", "disease_protein", "disease_disease", "drug_effect", "drug_protein"]

relations = {
    "phenotype present": "[ent1] has the phenotype [ent2]",
    "carrier": "[ent1] interacts with the carrier [ent2]",
    "enzyme": "[ent1] interacts with the enzyme [ent2]",
    "target": "The target of [ent1] is [ent2]",
    "transporter": "[ent2] transports [ent1]",
    "associated with": "[ent2] is associated with [ent1]",
    "parent-child": "[ent2] is a subclass of [ent1]",
    "side effect": "[ent1] has the side effect of [ent2]"
}
\end{lstlisting}

\section{Details for Hypergraph Transformer}
\label{apd:hygt}
First of all, we construct a hypergraph $\cG=(\cV,\cE)$ based on EHR data, where each patient visit is represented as a hyperedge connecting to all medical codes associated with the visit as nodes.
Then we utilize HyGT~\citep{cai2022hypergraph} to jointly learn the node and hyperedge embeddings. Specifically, The \textit{hyperedge embeddings} aggregate information from nodes within each hyperedge, while the \textit{node embeddings} aggregate information from hyperedges connecting the nodes.
In the $l$-th neural network layer, the node and hyperedge embeddings are updated as
\begin{equation}
\ \bX_{v}^{(l)} = f_{\mathcal{E} \rightarrow \mathcal{V}}\left(\cE_{v, \bE^{(l-1)}}\right),
\label{eq:node-update}
\end{equation}
\begin{equation}
\bE_{e}^{(l)} = f_{\mathcal{V} \rightarrow \mathcal{E}}\left(\cV_{e, \bX^{(l-1)}}\right),
\label{eq:hyperedge-update}
\end{equation}
where $\bX_{v}^{(l)}$ and $\bE_{e}^{(l)}$ represent the embeddings of node $v$ and hyperedge $e$ in the $l$-th layer ($1\leq l \leq L$), respectively. $\cE_{v, \bE}$ denotes the hidden representations of hyperedges that connect the node $v$, while $\cV_{e, \bX}$ is the hidden representations of nodes that are contained in the hyperedge $e$. 
The two message-passing functions $f_{\mathcal{V} \rightarrow \mathcal{E}}(\cdot)$ and $f_{\mathcal{E} \rightarrow \mathcal{V}}(\cdot)$ utilize multi-head self-attention  \citep{NIPS2017_3f5ee243} to identify significant neighbors during propagation as
\begin{equation}
\setlength{\abovedisplayskip}{9pt}
\setlength{\belowdisplayskip}{9pt}
f_{\mathcal{V} \rightarrow \mathcal{E}}(\bS)=f_{\mathcal{E} \rightarrow \mathcal{V}}(\bS)=\operatorname{Self-Att}(\bS), 
\nonumber
\end{equation}
where $\bm{S}$ is the input embedding for the attention layer, 
$\text{Self-Att}(\bm{S}) = \text{LayerNorm}(\bm{Y} + \text{FFN}(\bm{Y}))$. 
\( \bm{Y} \) is the output from the multi-head self-attention block
$\bm{Y} = \text{LayerNorm}(\bm{S} + \big\|_{i=1}^{h} \text{SA}_i(\bm{S}))$, 
\( \text{SA}_i(\bm{S}) \) denotes the scaled dot-product attention:
\begin{equation}
\setlength{\abovedisplayskip}{9pt}
\setlength{\belowdisplayskip}{9pt}
\text{SA}_i(\bm{S}) = \text{softmax}\left( \frac{\bm{W}_i^Q(\bm{SW}_i^K)^\top}{\sqrt{\lfloor d/h \rfloor}} \right)\bm{SW}_i^V.
\label{eq:scaled_dot_product}
\nonumber
\end{equation}
\( \bm{W}_i^Q \), \( \bm{W}_i^K \), and \( \bm{W}_i^V \) are learnable parameters for the \( i \)-th head corresponding to queries, keys, and values, respectively. 
To interpret the above process, the input sequence \( \bm{S} \) is projected into different \( h \) heads. The output of each head is then concatenated (denoted by \( \big\| \)) to form the multi-head attention output. This output of multi-head attention layer $\bm{Y}$ is then fed into a feed-forward neural network (FFN), comprising a two-layer Multilayer Perceptron (MLP) with ReLU activation functions.

\section{Details for Prompt Design}
\label{apd:prompt}
We present the detailed design of the prompt template as follows:
% Suppose you are a physician working on a health-related outcome prediction task and need to get relevant information for the given <domain>. Here are some relevant information:
% <domain> Name: <disease\_name>
% Retrieve Passage #1: 
% ...
% \n Based on the above information, Could you generate 1 sentence around 10-20 words to summarize the knowledge for the <domain> that is useful for <task\_name>?
\vspace{1ex}
\begin{Verbatim}
Suppose you are a physician working on a health
    -related outcome prediction task and need 
    to get relevant information for the given 
    <task>. Here is relevant information:
<medical code type> Name: <medical code name>
Retrieve Passage #1: <retrieved document 1>
Retrieve Passage #2: <retrieved document 2>

...
Based on the above information, Could you 
    generate 1 sentence of around 10-20 words 
    to summarize the knowledge for the 
    <medical code type> that is useful for
    the <task>?
\end{Verbatim}
<task> is the brief description of the downstream task. <medical code type> is either ``disease'', ``medication'' or ``procedure'', depending on the input.

\section{Human Study Guidelines}
\label{apd:additional}
% \begin{figure*}[h]
%     \centering    \includegraphics[width=0.99\linewidth]{latex/figures/ram-ehr-overview.pdf}
%     \vspace{-2ex}
%     \caption{An overview of {\ours}.\vspace{-2ex}}
%     % \vspace{-ex}
%     \label{fig:overview}
% \end{figure*}

For the human studies in Section~\ref{sec:add}, we provide the following guidelines for the annotators to evaluate the quality of the generated knowledge.
% \vspace{-2ex}
\begin{lstlisting}[linewidth=\columnwidth,breaklines=true,breakindent=0pt,basicstyle=\footnotesize\ttfamily]
The goal of this evaluation is to assess the helpfulness of generated knowledge explaining or relating to specific medical codes in the context of target prediction tasks. Helpfulness is defined by the relevance, accuracy, and utility of the information in facilitating understanding or decision-making related to medical coding and its implications for predictive tasks.

Please rate the following generated knowledge with score 0, 1 or 2.

> 0: Irrelevant
Definition: The knowledge does not provide any relevant information related to the medical code in question. It might be factually accurate but completely off-topic or not applicable to the context of target prediction tasks.

> 1: Partially Relevant and Useful
Definition: The knowledge provides some relevant information but either lacks completeness, specificity, or direct applicability to target prediction tasks. It might include general facts or insights that are related to the medical code but does not fully support decision-making or understanding in a predictive context.

> 2: Very Useful
Definition: The knowledge directly addresses the medical code with accurate, relevant, and comprehensive information that is highly applicable to target prediction tasks. It should provide detailed understanding, or specific examples that facilitate decision-making, understanding, or application in predictive modeling.
\end{lstlisting}

\section{Cost Information}
\label{apd:cost}
Utilizing \texttt{GPT-3.5-turbo} as our base LLM model for generating summarized knowledge, we observe an average retrieval augmentation cost of \$0.0025 per medical code in MIMIC-III and \$0.0032 in \dataset{}. Consequently, \ours{} does not result in excessive monetary expenses.

\end{document}